\documentclass{article}
\usepackage{ijcai25}

\usepackage{times}
\usepackage{soul}
\usepackage{url}
\usepackage[hidelinks]{hyperref}
\usepackage[utf8]{inputenc}
\usepackage[small]{caption}
\usepackage{graphicx}
\usepackage{amsmath}
\usepackage{amsthm}
\usepackage{booktabs}
\usepackage{algorithm}
\usepackage{algorithmic}
\usepackage[switch]{lineno}

\usepackage{subfigure}
\usepackage{amsmath}
\usepackage{amsthm}
\usepackage{booktabs}
\usepackage{algorithm}
\usepackage{algorithmic}
\usepackage[switch]{lineno}
\usepackage{multirow}
\usepackage{amsfonts,amssymb}
\usepackage{svg}

\title{Active Multimodal Distillation for Few-shot Action Recognition}

\author{
Weijia Feng$^1$
\and
Yichen Zhu$^1$\and
Ruojia Zhang$^1$\and
Chenyang Wang$^{2,3}$\footnote{Corresponding author: Chenyang Wang.}\and \\
Fei Ma$^3$\and 
Xiaobao Wang$^4$\And
Xiaobai Li$^{5,}$$^6$\\
\affiliations
$^1$College of Computer and Information Engineering, Tianjin Normal University, Tianjin, China\\
$^2$College of Computer Science and Software Engineering, Shenzhen University, Shenzhen, China\\
$^3$Guangdong Laboratory of Artificial Intelligence and Digital Economy (SZ), Shenzhen, China\\
$^4$College of Intelligence and Computing, Tianjin University, Tianjin, China\\
$^5$The State Key Laboratory of Blockchain and Data Security, Zhejiang University, Hangzhou, China\\
$^6$Hangzhou High-Tech Zone (Binjiang) Institute of Blockchain and Data Security, Hangzhou, China\\
\emails
weijiafeng@tjnu.edu.cn,
\{zhuyiiichen, zrj20001127\}@163.com,
chenyangwang@ieee.org,
mafei@gml.ac.cn,
wangxiaobao@tju.edu.cn,
xiaobai.li@zju.edu.cn
}

\begin{document}

\maketitle

\begin{abstract}
Owing to its rapid progress and broad application prospects, few-shot action recognition has attracted considerable interest. However, current methods are predominantly based on limited single-modal data, which does not fully exploit the potential of multimodal information. This paper presents a novel framework that actively identifies reliable modalities for each sample using task-specific contextual cues, thus significantly improving recognition performance. Our framework integrates an Active Sample Inference (ASI) module, which utilizes active inference to predict reliable modalities based on posterior distributions and subsequently organizes them accordingly. Unlike reinforcement learning, active inference replaces rewards with evidence-based preferences, making more stable predictions. Additionally, we introduce an active mutual distillation module that enhances the representation learning of less reliable modalities by transferring knowledge from more reliable ones. Adaptive multimodal inference is employed during the meta-test to assign higher weights to reliable modalities. Extensive experiments across multiple benchmarks demonstrate that our method significantly outperforms existing approaches.
\end{abstract}
\section{Introduction}

Over the past few years, video action recognition has witnessed substantial advancements, largely due to the rapid development of deep learning technologies. Although traditional approaches, including 2D and 3D Convolutional Neural Networks (CNNs), have shown remarkable proficiency in capturing both spatial and temporal features within videos, they typically demand extensive labeled data for effective training. The acquisition of such large-scale labeled datasets is often costly and time-intensive, especially for practical action recognition applications. To address this issue, recent research~\cite {yang2023aim} has focused on enhancing the efficiency of video understanding through lightweight models that can process temporal information efficiently while minimizing few-shot action recognition costs.

The existing methods~\cite{feng2024exploring,wang2024dependency} have shown significant performance. They mainly rely on single-modal data (such as RGB frames), and human actions comprise modalities such as RGB, optical flow, skeletal nodes, and depth information. These methods cannot capture the global information of human actions, and their performance will decrease to a certain extent when encountering complex actions. Due to the limitations inherent in single-modality approaches~\cite{ma2024review}, the potential of multiple modalities has attracted extensive scholarly attention and been empirically validated as effective~\cite{xue2024human,ma2024generative}. Several state-of-the-art methods have begun to combine multiple modalities~\cite{wu2025enriching,wang2025elevating,ma2024multi}, such as visual data, optical flow, and audio, to provide complementary recognition. However, these methods cannot identify which modality is important and which other modalities are not important in the current sample. These methods may assign extremely high weights to unimportant modalities when recognizing actions, resulting in unsatisfactory recognition results.

Inspired by Active Inference~\cite{tschantz2020reinforcement}, this paper proposes an Active Multimodal Few-Shot Inference for Action Recognition (AMFIR) to address the limitations of single-modal data in few-shot action recognition. The proposed AMFIR significantly improves the accuracy and efficiency of inference by actively identifying the most dominant modality of each query sample. 
This framework adopts a meta-learning paradigm, where each learning unit consists of labeled support samples and unlabeled query samples. 
In the meta-training phase, we use a modality-specific backbone network to extract feature representations based on active inference, and divide the query samples into an RGB dominant group and an optical-flow dominant group. 
We further designed a bidirectional distillation mechanism to guide the learning of unreliable modes through reliable modes. In the meta-test phase, Active Multimodal Inference (AMI) dynamically fuses posterior distributions of different modalities, assigning higher weights to more reliable modalities to optimize inference results.
Overall, the main contributions of this article can be summarized as follows:
\begin{itemize}
    \item This article utilizes the natural complementarity between different modalities to select the most dominant modality for each query sample through active inference, thereby significantly improving the performance of few-shot action recognition.
    \item A mutual refinement strategy has been proposed to transfer task-related knowledge learned from reliable modes to representation learning of unreliable modalities, leveraging the complementarity of multiple modalities to enhance the ability to identify unreliable modalities.
    \item We have designed an adaptive multimodal few-sample inference method that combines the results of specific modalities and assigns higher weights to more reliable modalities to optimize recognition performance further.
\end{itemize}

\begin{figure*}[h]
\centering
\includegraphics[width=0.93\linewidth]{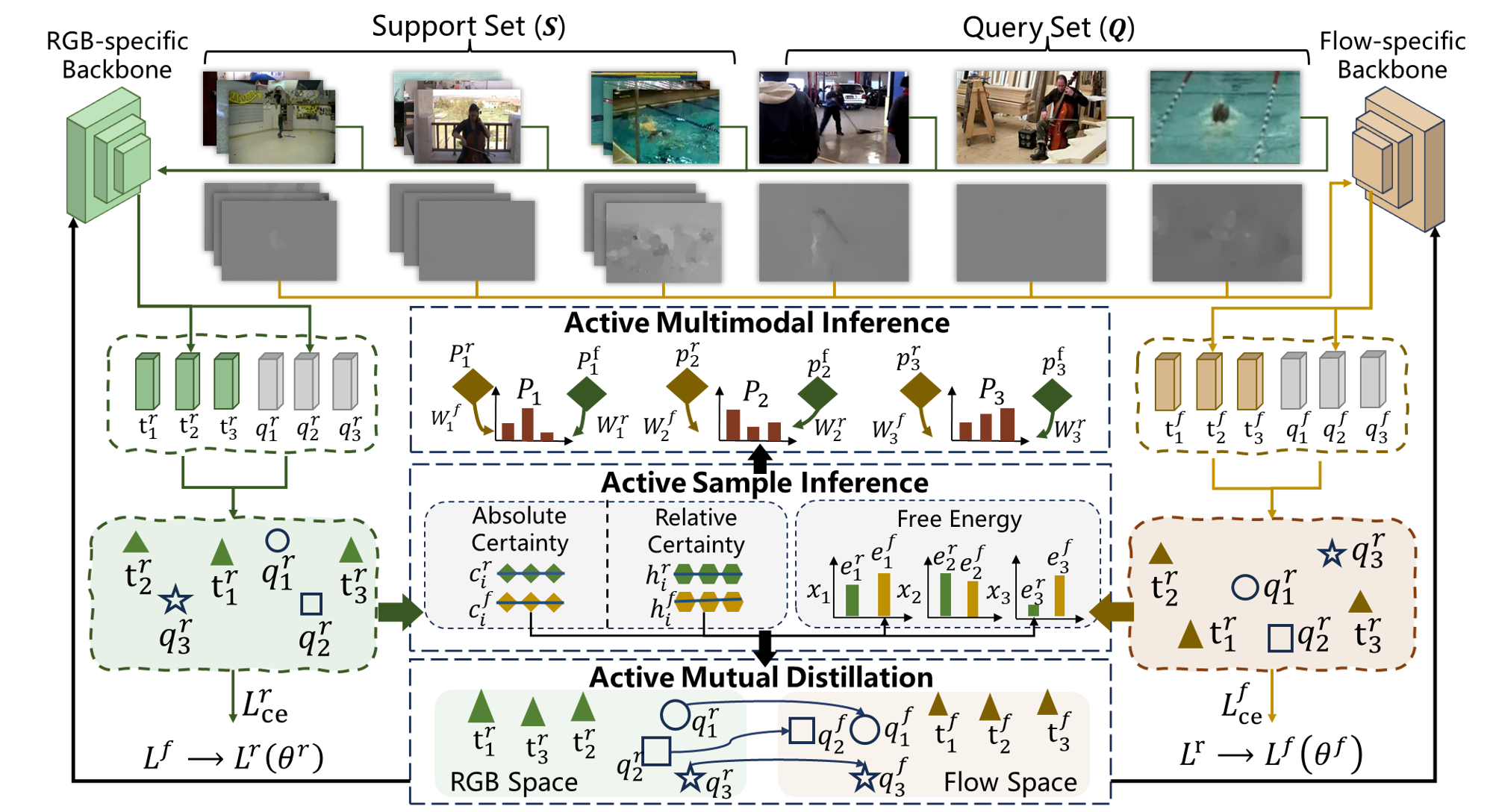}
\caption{Illustration of our proposed framework in the 3-way 3-shot setting.}
\label{fig:framework}
\end{figure*}

\section{Related Work}

\subsection{Few-shot Learning}

Few-shot learning (FSL) focuses on recognizing novel concepts with minimal labeled training data. In recent years, few-shot learning (FSL) has made significant strides in various traditional domains, such as image classification~\cite{bateni2020improved}, object detection~\cite{fan2020few,ma2024multi}, and segmentation~\cite{liu2020crnet,feng2025federated}. Despite these achievements, most existing FSL methods primarily concentrate on single-modality data, with relatively limited exploration of multimodal approaches. For instance, some techniques~\cite{pahde2019self} enhance low-shot visual embeddings by incorporating auxiliary text data during training, thereby boosting performance in few-shot image classification. Others~\cite{dong2018fast} focus on modeling the interplay between visual and textual information to address tasks like few-shot image description and visual question answering. Similarly, ~\cite{tsimpoukelli2021multimodal} leverages pre-trained language models to extend few-shot learning capabilities to downstream multimodal tasks, including visual question answering. Collectively, these studies highlight the potential of multimodal data to address the limitations inherent in unimodal FSL methods.

\subsection{Few-shot Action Recognition}

Two promising few-shot action recognition techniques have been proposed in the study. The first type utilizes data augmentation to support robust representation learning by creating supplementary training data~\cite{kumar2019protogan}, self-supervised cues~\cite{zhang2020few}, or auxiliary information~\cite{fu2020depth,wu2022motion}. Premier TACO~\cite {zheng2024premier} has improved the efficiency of the few-shot learning strategy through multi-task feature representation and negative sample selection mechanism, demonstrating significant performance. The second method focuses on alignment and strives to evaluate the similarity between query samples and support samples through synchronization frames or periods in the temporal or spatial domain~\cite{cao2020few,wang2022hybrid,wu2022motion}. The M2-CLIP framework~\cite{wang2024multimodal} utilizes multimodal adapters and multitasking decoders to improve video action recognition while maintaining a strong zero sample generalization ability.

\subsection{Active Inference}

Active Inference (AIF) is a Bayesian framework explaining how organisms minimize uncertainty and surprise by predicting and evaluating sensory inputs. It originates from Karl Friston’s Free Energy Principle (FEP) and emphasizes how organisms predict future states and adjust behavior to achieve goals through generative models during environmental interactions. For example, ~\cite{sedlak2024adaptive} explored the application of active reasoning in edge computing, demonstrating its potential in adaptive flow processing to meet service level objectives (SLOs) and real-time system management. Similarly, ~\cite{pezzulo2023generating} highlighted the close relationship between active reasoning and Generative AI, emphasizing its role in achieving higher-level intelligence and understanding through active perception and action. These advancements suggest that active reasoning provides a novel perspective for designing and implementing artificial intelligence, fostering progress in related fields. Furthermore, in hyperspectral image classification (HSIC), the Active Inference Transfer Convolutional Fusion Network (AI-TFNet) proposed by~\cite{wang2023ai} utilizes a pseudo-label propagation algorithm to enhance the availability of training samples and optimize classification performance.

\subsection{Knowledge Distillation}

Knowledge distillation serves as an effective knowledge transfer technique, extracting information from teacher networks and conveying it to student models. Recent research~\cite{hinton2015distilling,park2019relational,tung2019similarity} has highlighted its potential in cross-modal tasks. For instance, ~\cite{gupta2016cross} developed a method to transfer supervision across different modalities, using representations learned from well-labeled modalities as guidance for training new, unlabeled modalities. Similarly, ~\cite{garcia2018modality} proposed a technique for training multimodal video action recognition models using both depth and RGB data, addressing challenges such as noise and missing modalities during tests. The MARS framework~\cite{crasto2019mars} simulates optical flow through RGB frame training, avoiding optical flow computation at test time. It integrates appearance and motion information by combining feature loss and cross-entropy loss. Additionally, ~\cite{dai2021learning} introduced a knowledge distillation framework for action detection, enhancing RGB representations by leveraging knowledge from other modalities like optical flow and 3D pose. This approach achieves performance comparable to dual-stream networks using only RGB data during the test.

\section{Method}
\subsection{Problem Definition}

This paper employs a meta-learning framework for few-shot multimodal action recognition, which includes two primary phases: meta-training and meta-test. In the meta-training phase, we utilize a multimodal video dataset $D_{train}$ that encompasses base action classes $C_{train}$. We construct multiple meta-tasks (often termed episodes) from $D_{train}$ to train a meta-learner capable of generalizing to new action classes.  Each meta-task $\tau$ consists of a query set $Q \subset D_{train}$ and a support set $S \subset D_{train}$. Within the $N-way$ $K-shot$ meta-learning setting, the query set $Q = \{(x_i^r, x_i^f, y_i)\}_{i=1}^M$ includes $M$ multimodal query samples. Here, $x_i^r$ and $x_i^f$ represent two modalities (RGB and optical flow) of the $i-th$ query sample, and $y_i \in \{1,2,\cdots, N\}$ denotes the class label of the $i-th$ query sample. The support set $S = \{(x_i^r, x_i^f, y_i)\}_{i=M+1}^{M+NK}$ contains $K$ multimodal samples for each of the $N$ classes. In the meta-test phase, we employ a multimodal dataset $D_{test}$, which includes novel action classes $C_{test}$  that are disjoint from the training classes ($C_{test} \cap C_{train} = \oslash$). Similar to the meta-training phase, the support and query sets for each test task are constructed in the same manner. A key point to highlight is that the class labels of the query samples are concealed during the meta-test. The meta-learner must accurately classify each sample in the query set, relying exclusively on the labeled samples provided in the support set.

\subsection{Overview}

The overall architecture of the proposed AFMIR is depicted in Figure~\ref{fig:framework}. 
For each episode, we utilize a backbone network $\phi^m(Q,S;\theta^m), m \in (r,f)$ to extract the feature representations of query samples $\{q_i^m\}_{i=1}^M$ and  the prototypes of support samples $\{t_i^m\}_{k=1}^N$ for each modality. Subsequently, we calculate the modality-specific posterior distributions for each query sample based on the distances between query samples and prototypes in the modality-specific feature space. During the meta-training phase, we introduce an Active Sample Inference (ASI) module, which takes the modality-specific posterior distributions as inputs and performs reliability inference to assess the reliability of each modality (RGB and optical flow) for each query sample. This process categorizes the samples into two groups: the optical flow dominant group $\mathcal{G}^f$ and RGB dominant group $\mathcal{G}^r$. For the selected samples in $\mathcal{G}^f$ and $\mathcal{G}^r$, we implement Active Mutual Distillation to transfer task-specific knowledge from reliable modalities to unreliable ones via a bidirectional distillation mechanism, thereby enhancing the representation learning of the less dominant modality. In the meta-test phase, we employ adaptive multimodal inference, which leverages modality-specific posterior distributions to make adaptive fusion decisions, prioritizing the more reliable modalities.

\subsection{Active Sample Inference}

In this module, we use the idea of AIF to actively predict the most dominant modality of each sample, and this predicted modality will be considered as the sample-specific dominant modality of the corresponding sample. 
Before introducing this module, we first need to review the method of AIF, which aims to maximize the Bayesian model evidence for an agent's generative model in the context of Partially Observed Markov Decision Processes (POMDP). 
Formally, a POMDP is defined by a tuple $\left \langle \mathcal{S}, \mathcal{A}, \mathcal{O}, \mathcal{P}, \Theta \right \rangle$, where $\mathcal{S}$ denotes the true state of the environment and $\mathcal{A}$ denote agent's action space. 
During each time step $t$, the agent transitions to a new state $s_t$, which is calculated through $P(s_t | s_{t-1}, a_{t-1})$, where the transition is informed by ${s_{t-1} \in \mathcal{S}}$ and ${a_{t-1} \in \mathcal{A}}$ executed at the preceding time step $t-1$.
Agents might not have direct access to the actual state of the environment. Instead, they receive observations, denoted as $ o_t $, sampled from a distribution $ P(o_t \mid s_t) $ that depends on the true environmental state $ s_t $. 
Given this limitation, agents must base their operations on an inferred belief about the true state, $ \hat{s}_t $, representing their estimation of $ s_t $ based on the received observations.
Within the framework of a specified generative model, agents perform approximate Bayesian inference through the encoding of an arbitrary probability distribution $q(s,\theta)$, which is optimized through the minimization of variational free energy $\tilde{\mathcal{F}}$:
\begin{equation} 
\label{free energy eq}
    \begin{aligned}
        \tilde{\mathcal{F}} = D_{KL}\left(q(s, \theta)\|p^\Phi(o, s, \theta)\right),
    \end{aligned}
\end{equation}
where $o \in \mathcal{O} $ denotes the agent's observations and $\theta \in \Theta$ represents model parameters.

To establish the active inference model, we input the representations of query samples $\{q_i^m\}_{i=1}^M$ and the prototypes of support samples $\{t_k^m\}_{k=1}^N$ into the module. We consider the dominant modality for each query sample as the one that can reflect more task-specific discriminative features. However, the dominant modality for each query sample is not fixed, as the contribution of a specific modality largely depends on the contextual information of the query and support samples in each task. To address this, we propose to infer the reliability of different modalities based on modality-specific posterior distributions. This approach aligns with the core principles of active inference, where the model dynamically assesses the reliability of different modalities to optimize its predictive capabilities.
For each query sample, the modality-specific posterior distribution $p_i^m$ can be formulated as:
\begin{equation}
p_i^m(k | x_i^m) = \frac{e^{-\psi(q_i^m, t_k^m)}}{\sum_{k'=1}^{N} e^{-\psi(q_i^m, t_{k'}^m)}}, 
\label{eq:posterior}
\end{equation}
where $k \in \{1, \dots, N\}, m \in \{r, f\}$, and $q_i^m$ denotes the modality-specific representation of the $i$-th query sample, $t_k^m$ denotes the prototype of the $k$-th class, and $\psi$ is a distance measurement function (\emph{e.g.,} Euclidean distance in this work). 
The posterior distribution reflects the model's belief in the class labels given the specific modality, consistent with minimizing prediction errors to optimize beliefs. 
By dynamically evaluating the reliability of each modality in the context of the current task, the model can select the most informative modality to reduce prediction errors.

After obtaining the modal-specific posterior distribution $p_i^m$ for the $i^{th}$ query sample, we use it as input to construct the observation space $\mathcal{O}$ and the state space $\mathcal{S}$.
The reliability of each modality for a given sample can be assessed by the free energy derived from Eq. (\ref{free energy eq}).
To elaborate, the modality with the lower free energy value would be considered more reliable according to the calculations based on the provided formula. 
This approach enables a quantitative comparison between different modalities, facilitating a more accurate determination of their respective reliabilities.
In order to systematically investigate the cross-modal complementarity, a selection criterion is established based on the significant discrepancy in reliability between the two modalities. Consequently, the queried samples are meticulously categorized into the following distinct groups:
\begin{equation}
    \mathcal{G}^m = \{ (x^r_i, x^f_i) \mid (x^r_i, x^f_i) \in \mathcal{G}, \; \mathcal{F}^m_i > \mathcal{F}^n_i \},
\end{equation}
where $m, n \in \{r, f\}, \; m \neq n$, and $\mathcal{G}^m$ denotes the group dominated by modality $m$, comprising query samples for which the modality  $m$ exhibits higher certainty than modality $n$ in the few-shot task. Specifically, when $m = f$,  $\mathcal{G}^m$ represents the Flow-dominant group. 
Conversely, when $m = r$,  $\mathcal{G}^m$ corresponds to the RGB-dominant group, indicating that the RGB modality is more reliable for distinguishing between query samples.

\subsection{Active Mutual Distillation}
In this section, we first introduce an active mutual distillation method, which enhances the representation learning of less reliable modes by utilizing task-specific discriminative knowledge from more reliable modes. Traditional knowledge extraction methods typically involve using well-trained teacher models to guide student model learning through consistency constraints, such as KL divergence calculated based on logits:
\begin{equation}
    D_{KL}(p^t \| p^s) = \sum_{i=1}^{N} p^t_i \left( \log p^t_i - \log p^s_i \right),
\end{equation}
where $p^t$ and $p^s$ represent the logits produced by the teacher and student models, respectively. In conventional methods, teacher-student distillation is consistently applied to individual samples.

Based on this, we define the absolute certainty $c_i^m$ as the maximum value of the modality-specific posterior distribution, which quantifies the model's highest confidence in any class prediction for modality $m$:
\begin{equation}
    c_i^m = \max_{k} p_i^m(k | x_i^m),
\label{equ:c}
\end{equation}
where $p_i^m(k | x_i^m)$ denotes the posterior distribution over classes for the $i$-th query sample in modality $m$. Absolute certainty $c_i^m$ reflects the highest belief of the model in its prediction for that modality. This dynamic assessment allows us to adaptively evaluate the reliability of different modalities based on contextual information in each task, thus better accommodating the specific demands of different tasks.

To exploit the complementarity between different modalities and improve few-shot action recognition, we refine the approach by treating models trained on one modality as teachers and those trained on another modality as students. However, determining which modality should act as the teacher is challenging, as the contribution of each modality varies between samples and depends heavily on the contextual information of the few-shot task. To address this, we previously introduced active inference to dynamically infer the importance of different modalities for each sample. We actively assign more reliable modalities as teachers to transfer knowledge. Specifically, we constrain the learning of two modality-specific models by actively transferring query-to-prototype relationship knowledge between different modalities:
\begin{equation}
\mathcal{L}_{m \rightarrow n}(\theta^n) = \frac{1}{\sum_{(x_i^n,x_i^m)}c_i^m}\sum_{(x_i^n,x_i^m)}c_i^m D_{KL}(p_i^m,p_i^n),
\end{equation}
where $p_i^m$ denotes the modality-specific posterior distribution for the $i$-th query sample in modality $m$, as defined in Eq.(\ref{eq:posterior}), $c_i^m$ represents the absolute certainty for the $i$-th query sample in modality $m$, as defined in Eq. (\ref{equ:c}), and $m, n \in \{r, f\}$ with $m \neq n$.

\subsection{Active Multimodal Inference}

In the first two sub-sections, we explored how different modalities of each sample contribute to meta-training and discussed their role in the meta-test phase. Specifically, we introduce a method for adaptively integrating multimodal predictions to form the final decision in few-shot inference. Given the varying reliability of modalities for each query sample, we design an adaptive multimodal fusion strategy as follows:
\begin{equation}
    \mathcal{P}(k|x_i^m) = \frac{e^{-\alpha_i^r \psi(q_i^r, t_k^r) - \alpha_i^f \psi(q_i^f, t_k^f)}}{\sum_{k'=1}^{N} e^{-\alpha_i^r \psi(q_i^r, t_{k'}^r) - \alpha_i^f \psi(q_i^f, t_{k'}^f)}},
\end{equation}
where $\psi(\cdot)$ represents the Euclidean distance function, and $\alpha_i^r$ and $\alpha_i^f$ are the adaptive fusion weights for the RGB and optical flow modalities of the $i$-th query sample, respectively.

Since modality-specific posterior distributions may not always be accurate during meta-tests, and relative certainty does not directly reflect the similarity between query samples and class prototypes, we use the absolute certainty values $c_i^m$ to compute the adaptive fusion weights:
\begin{equation}
    \alpha_i^m = \frac{c_i^m}{c_i^r + c_i^f},
\end{equation}
where $m \in \{r, f\}$ indicates the modality, where $r$ represents the RGB modality and $f$ represents the optical flow modality.

\subsection{Optimization}
The proposed AFMIR can be optimized with the following function:
\begin{equation}\small
    \mathcal{L} = \sum_{m \in \{r, f\}} \mathcal{L}^m_{ce}(\theta^m) + \lambda \sum_{m \neq n \in \{r, f\}} \mathcal{L}^m \rightarrow \mathcal{L}^n(\theta^n),
    \label{equ_loss}
\end{equation}
where $\mathcal{L}^m_{ce}(\theta^m)$ denotes the cross-entropy loss for modality $m$, and $\mathcal{L}^m \rightarrow \mathcal{L}^n(\theta^n)$ represents the knowledge distillation loss from modality $m$ to modality $n$. The parameter $\lambda$ balances the contribution of the distillation losses.
Besides, the $\mathcal{L}^m_{ce}(\theta^m)$ is further used to constrain the modality-specific predictions, \emph{i.e.,}
\begin{equation}
    \mathcal{L}^m_{ce}(\theta^m) = \sum_{i=1}^{M} \sum_{k=1}^{N} p^m_i(k) \log p^m_i(k),
\end{equation}
where $m \in \{r, f\}$ indicates the modality, $r$ represents the RGB modality and $f$ represents the optical flow modality. 
The cross-entropy loss $\mathcal{L}^m_{ce}(\theta^m)$ is computed for each modality based on the predicted probabilities $p^m_i(k)$ of the query samples belonging to each class $k$. 

The parameters of the modality-specific backbone networks $\theta^m, m \in \{r, f\}$ are optimized via distinct weighted combinations of their corresponding losses:
\begin{equation}\small
    \theta^m \leftarrow \theta^m - \gamma \nabla_{\theta^m} ( \mathcal{L}^m_{ce}(\theta^m) + \lambda \sum_{n \neq m \in \{r, f\}} \mathcal{L}^n \rightarrow \mathcal{L}^m(\theta^m)).
\end{equation}
The parameter update for each modality $\theta^m$ is performed by minimizing the cross-entropy loss $\mathcal{L}^m_{ce}(\theta^m)$ and the knowledge distillation loss from the other modality $n \neq m$. The learning rate is denoted by $\gamma$, and $\lambda$ balances the contribution of the distillation losses.

\section{Experiments}

\begin{table*}[h!]
\centering
\scalebox{1.1}{\begin{tabular}{c|c|cc|cc|cc|cc}
\hline
Modality   & Method                                                                                            & \multicolumn{2}{c|}{\begin{tabular}[c]{@{}c@{}}Kinetics\\1-shot 5-shot\end{tabular}}                                                                                  & \multicolumn{2}{c|}{\begin{tabular}[c]{@{}c@{}}SSv2\\ 1-shot 5-shot\end{tabular}}                                          & \multicolumn{2}{c|}{\begin{tabular}[c]{@{}c@{}}HMDB51\\ 1-shot 5-shot\end{tabular}}                                                                                      & \multicolumn{2}{c}{\begin{tabular}[c]{@{}c@{}}UCF101\\ 1-shot 5-shot\end{tabular}}                                                                                      \\ \hline
RGB        & \begin{tabular}[c]{@{}c@{}}Matching Net\\ TRX\\ HyRSM\\ STRM\end{tabular}                        & \multicolumn{1}{c}{\begin{tabular}[c]{@{}c@{}}53.3\\ 63.6\\ 73.7\\ -\end{tabular}}          & \begin{tabular}[c]{@{}c@{}}78.9\\ 85.9\\ 86.1\\ 86.7\end{tabular}       & \multicolumn{1}{c}{\begin{tabular}[c]{@{}c@{}}-\\ 42.0\\ 54.3\\ -\end{tabular}}              & \begin{tabular}[c]{@{}c@{}}-\\ 64.6\\ 69.0\\ 68.1\end{tabular}           & \multicolumn{1}{c}{\begin{tabular}[c]{@{}c@{}}-\\ -\\ 60.3\\ -\end{tabular}}              & \begin{tabular}[c]{@{}c@{}}-\\ 75.6\\ 78.0\\ 77.3\end{tabular}           & \multicolumn{1}{c}{\begin{tabular}[c]{@{}c@{}}-\\ -\\ 83.9\\ -\end{tabular}}              & \begin{tabular}[c]{@{}c@{}}-\\ 96.1\\ 94.7\\ 96.9\end{tabular}           \\ \hline
Flow       & \begin{tabular}[c]{@{}c@{}}ProtoNet-F\\ TRX-F\\ STRM-F\end{tabular}                               & \multicolumn{1}{c}{\begin{tabular}[c]{@{}c@{}}45.2\\ 44.8\\ 47.8\end{tabular}}              & \begin{tabular}[c]{@{}c@{}}69.5\\ 69.7\\ 69.7\end{tabular}              & \multicolumn{1}{c}{\begin{tabular}[c]{@{}c@{}}32.9\\ 30.7\\ 36.3\end{tabular}}               & \begin{tabular}[c]{@{}c@{}}51.1\\ 52.4\\ 55.7\end{tabular}               & \multicolumn{1}{c}{\begin{tabular}[c]{@{}c@{}}43.7\\ 43.0\\ 52.2\end{tabular}}               & \begin{tabular}[c]{@{}c@{}}65.0\\ 67.6\\ 67.9\end{tabular}               & \multicolumn{1}{c}{\begin{tabular}[c]{@{}c@{}}69.7\\ 65.6\\ 79.7\end{tabular}}               & \begin{tabular}[c]{@{}c@{}}89.6\\ 90.6\\ 91.6\end{tabular}               \\ \hline
Multimodal & \begin{tabular}[c]{@{}c@{}}ProtoNet-EC\\ ProtoNet-EA\\ STRM-EC\\ AFMAR\\ AFMIR(ours)\end{tabular} & \multicolumn{1}{c}{\begin{tabular}[c]{@{}c@{}}63.8\\ 61.7\\ 68.3\\ 80.1\\ \textbf{82.8}\end{tabular}} & \begin{tabular}[c]{@{}c@{}}84.1\\ 83.9\\ 87.4\\ 92.6\\ \textbf{96.1}\end{tabular} & \multicolumn{1}{c}{\begin{tabular}[c]{@{}c@{}}33.0\\ 31.1\\ 45.5\\ 61.7\\ \textbf{70.6}\end{tabular}} & \begin{tabular}[c]{@{}c@{}}49.5\\ 50.5\\ 66.7\\ 79.5\\ \textbf{92.3}\end{tabular} & \multicolumn{1}{c}{\begin{tabular}[c]{@{}c@{}}56.9\\ 53.2\\ 59.3\\ 73.9\\ \textbf{83.7}\end{tabular}} & \begin{tabular}[c]{@{}c@{}}73.8\\ 46.3\\ 78.3\\ 87.8\\ \textbf{90.0}\end{tabular} & \multicolumn{1}{c}{\begin{tabular}[c]{@{}c@{}}78.3\\ 76.7\\ 87.4\\ 91.2\\ \textbf{94.9}\end{tabular}} & \begin{tabular}[c]{@{}c@{}}93.9\\ 94.3\\ 96.3\\ 99.0\\ \textbf{99.1}\end{tabular} \\ \hline
\end{tabular}}
\caption{Comparison with state-of-the-art few-shot action recognition methods. The best results are in bold. For multi-modal methods extended from existing unimodal methods, ``EC'' represents cascaded early fusion schemes, ``EA'' represents collaborative attention early fusion. The `-' indicates that the result is not available in published works.} 
\label{tab:com}    
\end{table*}

\subsection{Validation Protocol}

\textbf{Datasets}. We assess our proposed method on four prominent and challenging benchmarks for few-shot action recognition: Kinetics-400~\cite{kay2017kinetics}, Something-Something V2~\cite{goyal2017something}, HMDB51~\cite{wang2015action}, and UCF101~\cite{peng2018two}. These datasets are augmented to include multi-modal data by generating optical flow frame sequences from the original videos using a dense optical flow algorithm.

\textbf{Modality-specific Backbones}. For the RGB modality, a pre-trained ResNet-50 backbone is employed to extract visual features at the frame level. For the optical flow modality, an I3D model pre-trained on the Charades dataset is used to capture motion features from individual frames. Subsequently, video-level features for both modalities are derived by aggregating these enhanced frame-level features. Ultimately, the query-specific prototype is generated by consolidating the video-level features from the support samples of the corresponding action class.

\textbf{Parameters}. In the meta-training phase, the balance weight ($\lambda$, specified in Eq.~\ref{equ_loss}) is uniformly set to 1.0 across all benchmarks. Training is conducted using the SGD optimizer. For both RGB and optical flow modalities, the respective networks are iteratively updated by minimizing a combined weighted loss function, which includes both cross-entropy and distillation losses, until convergence is achieved. 
The learning rate $\gamma$ is set as $10^{-3}$.


\begin{figure}[h!]
\centering
\subfigure[SSv2]{
\label{free_energy_sthv2}
\includegraphics[width=0.235\textwidth]{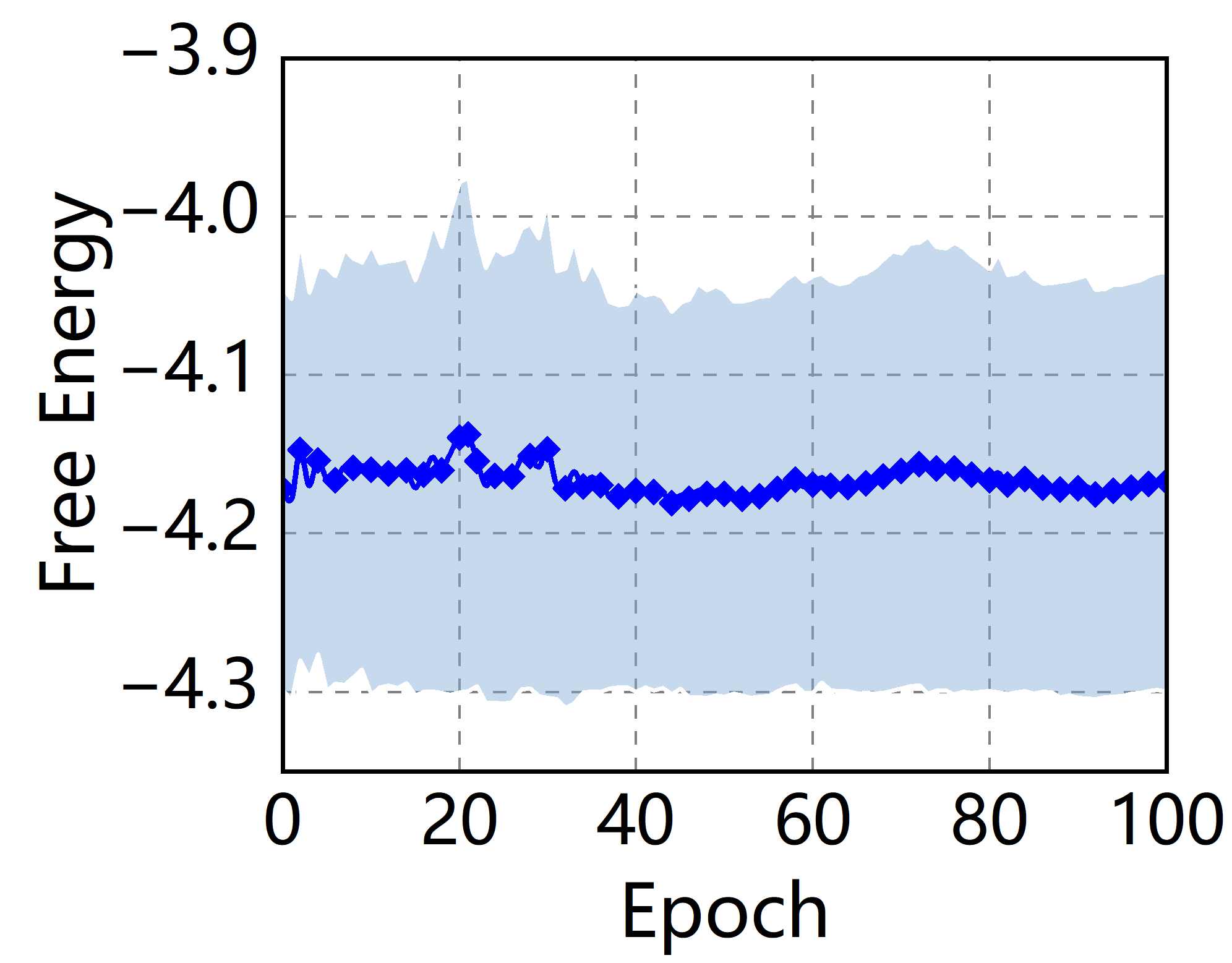}
}
\hspace{-0.15in}
\subfigure[Kinetics-400]{
\label{free_energy_kinetics}
\includegraphics[width=0.235\textwidth]{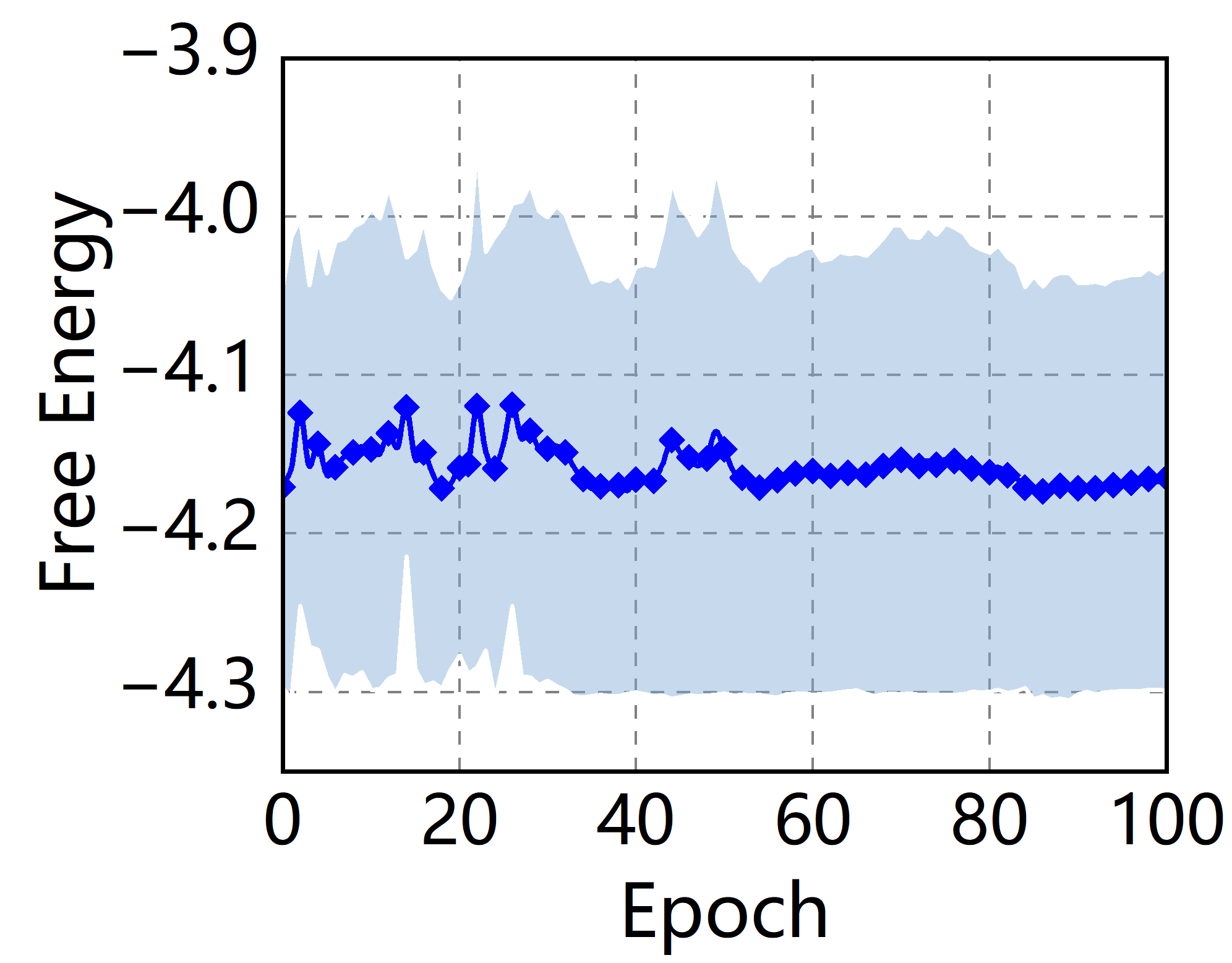}
}
\caption{The change of free energy.}
\label{free energy fig}
\end{figure}

\subsection{Free Energy}
The experimental results show that, on the SSv2 and Kinetics-400 datasets, there is a significant fluctuation in the initial free energy during training. However, as training progresses, the free energy gradually decreases and tends to stabilize, eventually converging to around -4.0 on both datasets (as shown in Figure~\ref{free energy fig}). This trend indicates that the model dynamically evaluates modal reliability through active inference modules, prioritizing the selection of modes with lower free energy for inference, which significantly optimizes the modal selection strategy. In addition, the knowledge distillation and dynamic weighting mechanisms between multiple modalities further reduce the uncertainty of modal selection, enhance task adaptation capability, and improve the stability of cross-modal inference. These results validate the crucial role of active inference and multimodal interaction in few-shot action recognition, enabling the model to more efficiently and reliably utilize modal information to complete tasks.

\subsection{Comparative Experiments}

We selected four RGB-based algorithms: Matching Net~\cite{zhu2018compound}, TRX~\cite{perrett2021temporal}, HyRSM~\cite{wang2022hybrid}, and STRM~\cite{thatipelli2022spatio}. These methods leverage metric learning, temporal relationships, spatiotemporal modeling, and hybrid relationships to advance few-shot action recognition. We retrained vision-based methods like TRX-F and STRM-F for optical flow data due to the lack of existing methods. We extended visual methods for multimodal baselines via early fusion (EC, EA) and late fusion (LF). Additionally, the chosen AFMAR~\cite{wanyan2023active} algorithm enhances few-shot recognition by actively selecting reliable modalities, distilling knowledge bidirectionally, and adaptively fusing multimodal data.

As shown in Table~\ref{tab:com}, our AMFIR framework significantly outperforms existing methods in Few Shot Action Recognition tasks on datasets such as SSv2, HMDB51, UCF101, and Kinetics-400. The accuracy of 1-shot and 5-shot tasks reaches 70.6\% and 92.3\% (SSv2), 83.7\% and 90.0\% (HMDB51), 94.9\% and 99.1\% (UCF101), and 82.8\% and 96.1\% (Kinetics-400), respectively. This outstanding performance is attributed to the active sample inference module (ASI) in the framework, which dynamically selects the most reliable mode to reduce uncertainty, the active mutual distillation module (AMD) improves the representation ability of unreliable modes through bidirectional knowledge distillation, and the Active Multimodal Inference module (AMI) optimizing the complementarity of modes through adaptive fusion, thus fully utilizing the potential of multimodal data.

\begin{table}[ht]
\centering
\scalebox{0.9}
{\begin{tabular}{cccc cc cc}
\hline
\multicolumn{2}{c}{ASI} & \multirow{2}{*}{AMD} & \multirow{2}{*}{AMI} & \multicolumn{2}{c}{Kinetics} & \multicolumn{2}{c}{SSv2} \\
RD & FD & & & 1-shot & 5-shot & 1-shot & 5-shot \\ 
\hline
$\times$ & \checkmark & \checkmark & \checkmark & 64.10 & 68.94 & 58.03 & 66.81 \\ 
\checkmark & $\times$ & \checkmark & \checkmark & 59.63 & 75.27 & 62.91 & 66.06 \\ 
\checkmark & \checkmark & $\times$ & \checkmark & 71.64 & 89.05 & 61.81 & 83.37 \\ 
\checkmark & \checkmark & \checkmark & $\times$ & 59.34 & 65.78 & 62.59 & 85.33 \\ 
\checkmark & \checkmark & \checkmark & \checkmark & \textbf{82.85} & \textbf{96.11}& \textbf{70.59} & \textbf{92.32} \\ 
\hline
\end{tabular}}
\caption{Comparison of results across different configurations. The best are in bold.}
\label{Tab-Abalation}
\end{table}

\subsection{Ablation Study}


\textbf{The impact of key components}.
The experimental results in Table~\ref{Tab-Abalation} have been validated through ablation studies, demonstrating that the complete framework has achieved state-of-the-art performance on both Kinetics and SSv2. When the ASI module uses only RGB as the dominant modality (RD) or only optical flow as the dominant modality, the accuracy will significantly decrease. Removing AMD will reduce the thermal accuracy of Kinetics 5 by 7.06\%. Compared with dynamics, disabling AMI resulted in a greater decrease in SSv2, indicating the necessity of time-sensitive adaptive fusion. 
\subsection{Further Remarks}

\begin{figure}[h!]
\centering
\subfigure[SSv2]{
\includegraphics[width=0.23\textwidth]{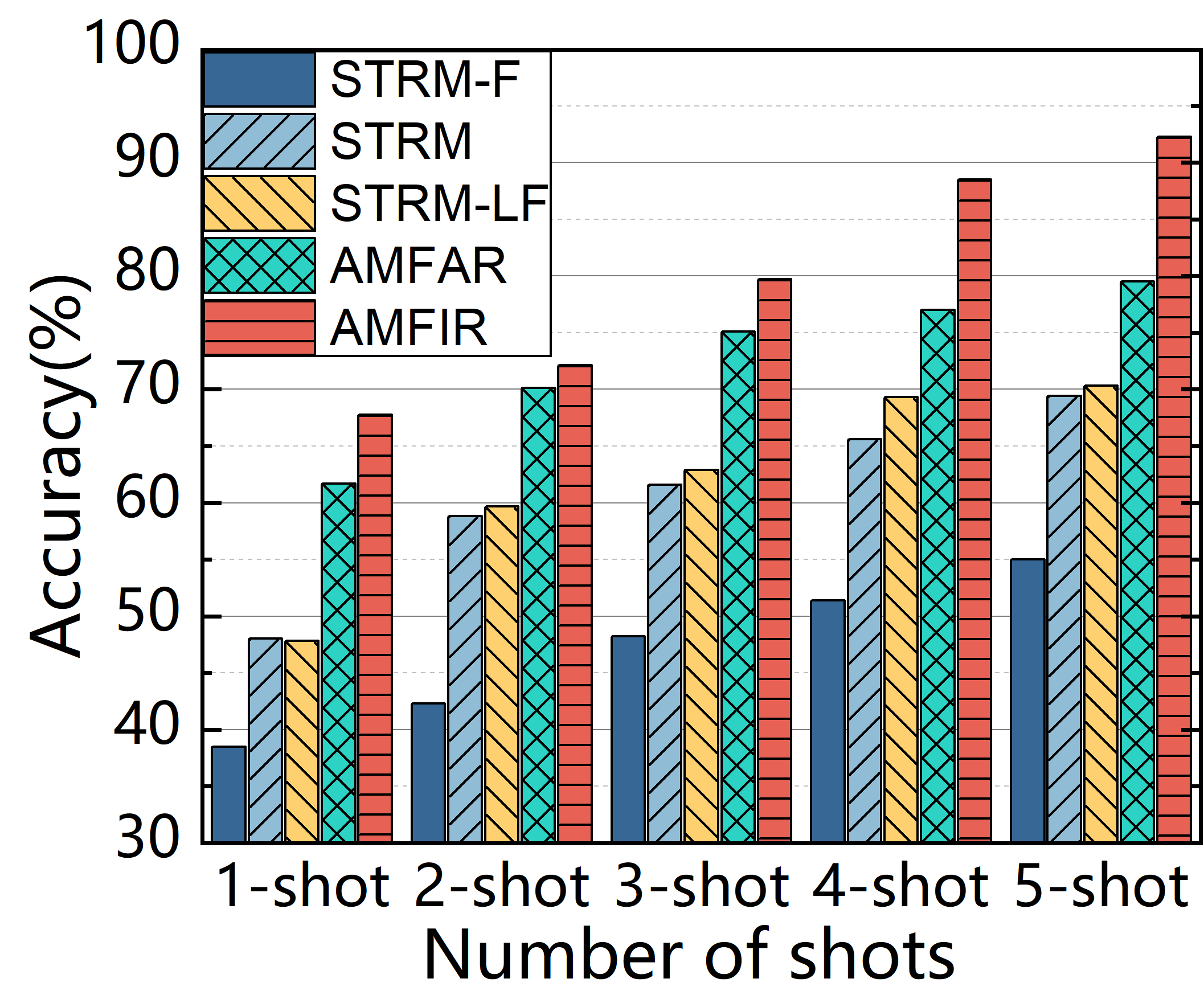}
}
\hspace{-0.1in}
\subfigure[Kinetics]{
\includegraphics[width=0.23\textwidth]{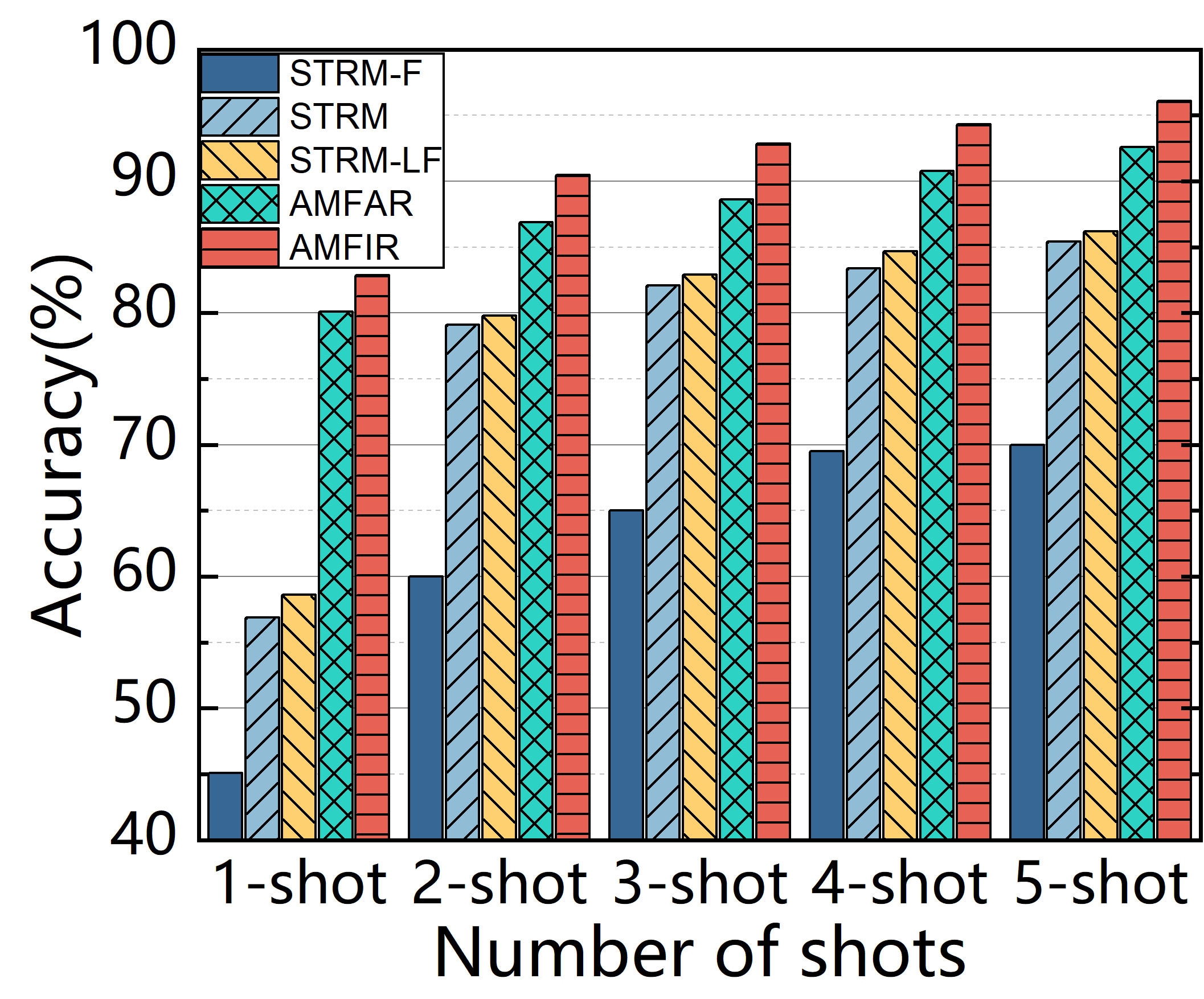}
}
\vspace{-0.3cm}
\caption{Comparison results with different numbers of support samples in 5-way K-shot setting.
}
\label{fig:5wayKshot}
\end{figure}

\textbf{Performance with Different Numbers of Support Samples.} The experimental results in Figure~\ref{fig:5wayKshot} show that the performance of AMFIR steadily improves as the number of support samples changes from 1-shot to 5-shot. The accuracy on the SSv2 and Kinetics datasets increases from about 70\% and 82\% to about 92\% and 96\%, respectively, which is significantly better than other methods, especially under few-shot conditions. The reason for these results is that the ASI module reduces uncertainty by dynamically selecting reliable modalities, enabling the model to have high initial performance at 1-shot. The AMD module utilizes bidirectional knowledge distillation to enhance the collaborative effect between modalities, supporting further performance improvement as the sample size increases. The AMI module adaptively adjusts modal weights and optimizes the integration strategy of multimodal data. The synergistic effect of these modules fully utilizes the complementarity of multimodal data, resulting in the excellent performance of AMFIR in both few-shot and multi-sample scenarios.

\vspace{-0.4cm}

\begin{figure}[h!]
\centering
\subfigure[SSv2]{
\label{Distillation_SSv2}
\includegraphics[width=0.23\textwidth]{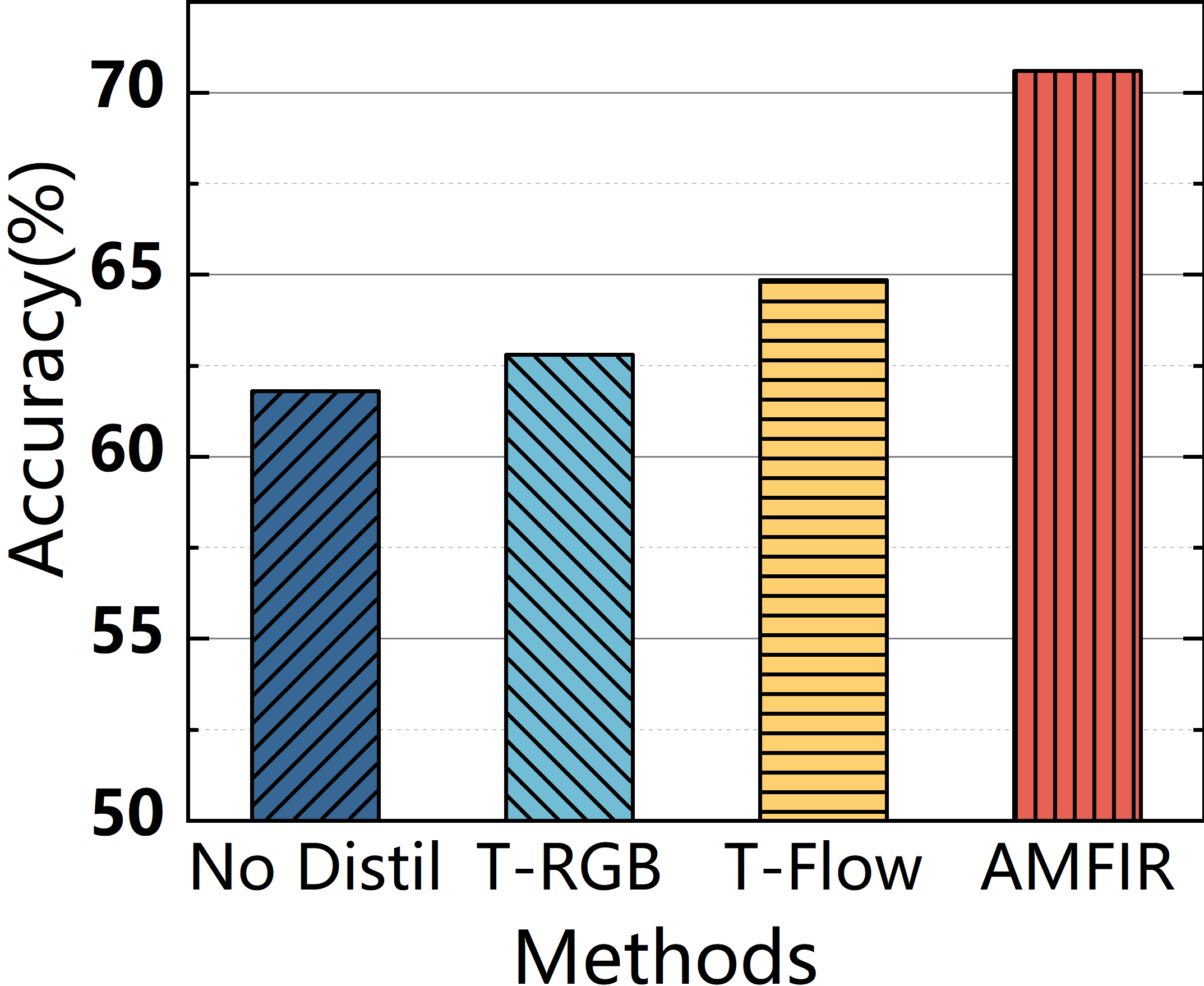}
}
\hspace{-0.1in}
\subfigure[Kinetics-400]{
\label{Distillation_Kinetics}
\includegraphics[width=0.23\textwidth]{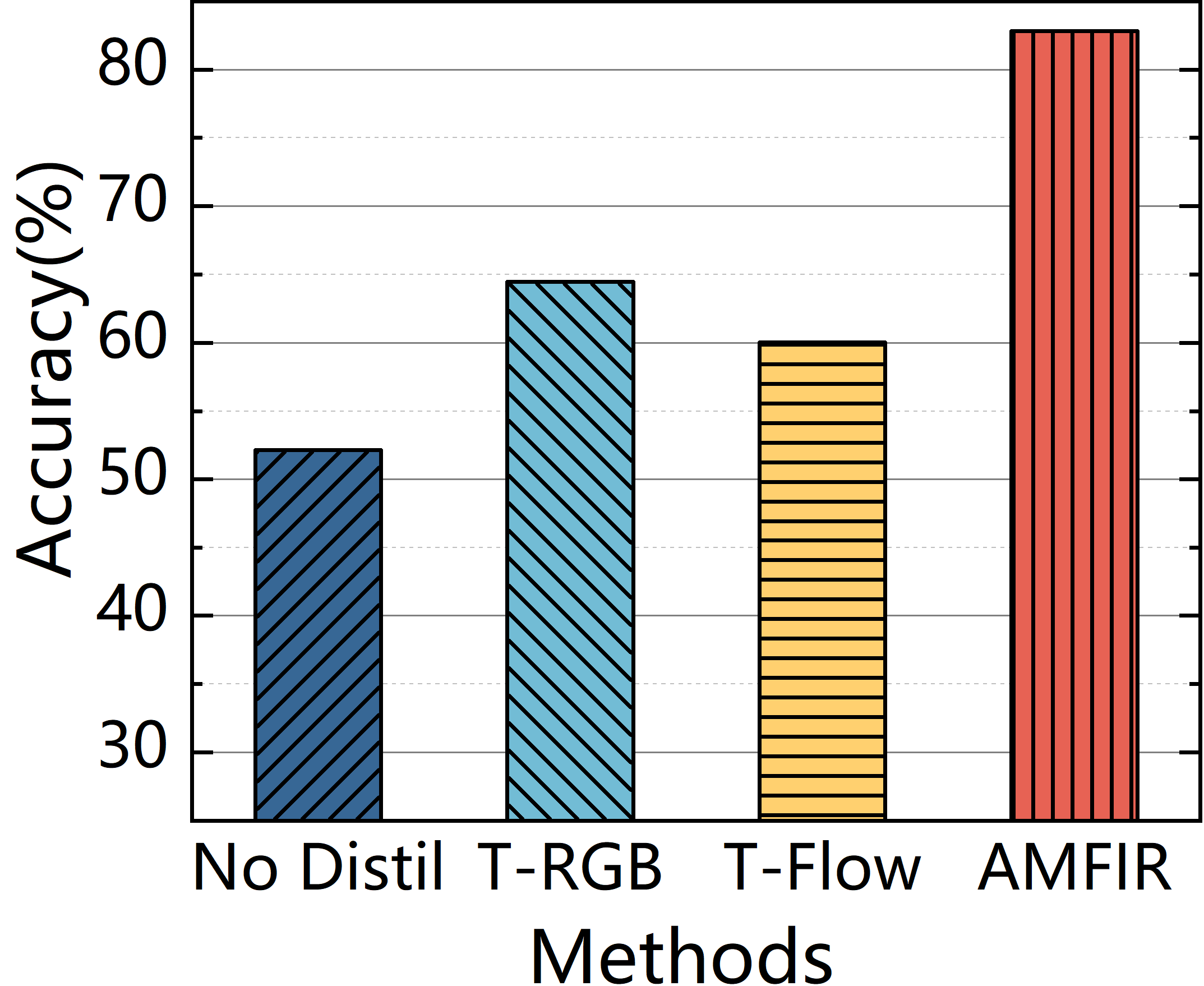}
}
\caption{Comparison with conventional distillation strategies in a 5-way 1-shot setting. T-RGB (or T-Flow) denotes distillation, whereas RGB (optical flow) is consistently regarded as the teacher.}
\label{Fig_Distillation}
\end{figure}

\textbf{Influence of Different Distillation Strategies.} The experimental results in Figure~\ref{Fig_Distillation} show that the bidirectional distillation strategy (AMFIR) achieved accuracies of 70.6\% and 82.8\% in the 5-way 1-shot task on the SSv2 and Kinetics-400 datasets, respectively, significantly outperforming the unidirectional distillation strategy (T-RGB and T-Flow) and the no-distillation strategy (No Distillation). The reason is that bidirectional distillation effectively transmits task-related knowledge of reliable modalities through the mechanism of mutual teaching, enhances the representation ability of unreliable modalities, and further optimizes the learning effect of reliable modalities. In addition, this strategy fully utilizes the complementarity between RGB and optical flow modalities and combines active inference to dynamically evaluate the modal reliability, flexibly optimizing the distillation process. This design demonstrates the significant advantages of bidirectional distillation in few-shot multimodal learning.

\begin{figure}[h!]

\vspace{-0.4cm}
\centering
\subfigure[SSv2]{
\includegraphics[width=0.23\textwidth]{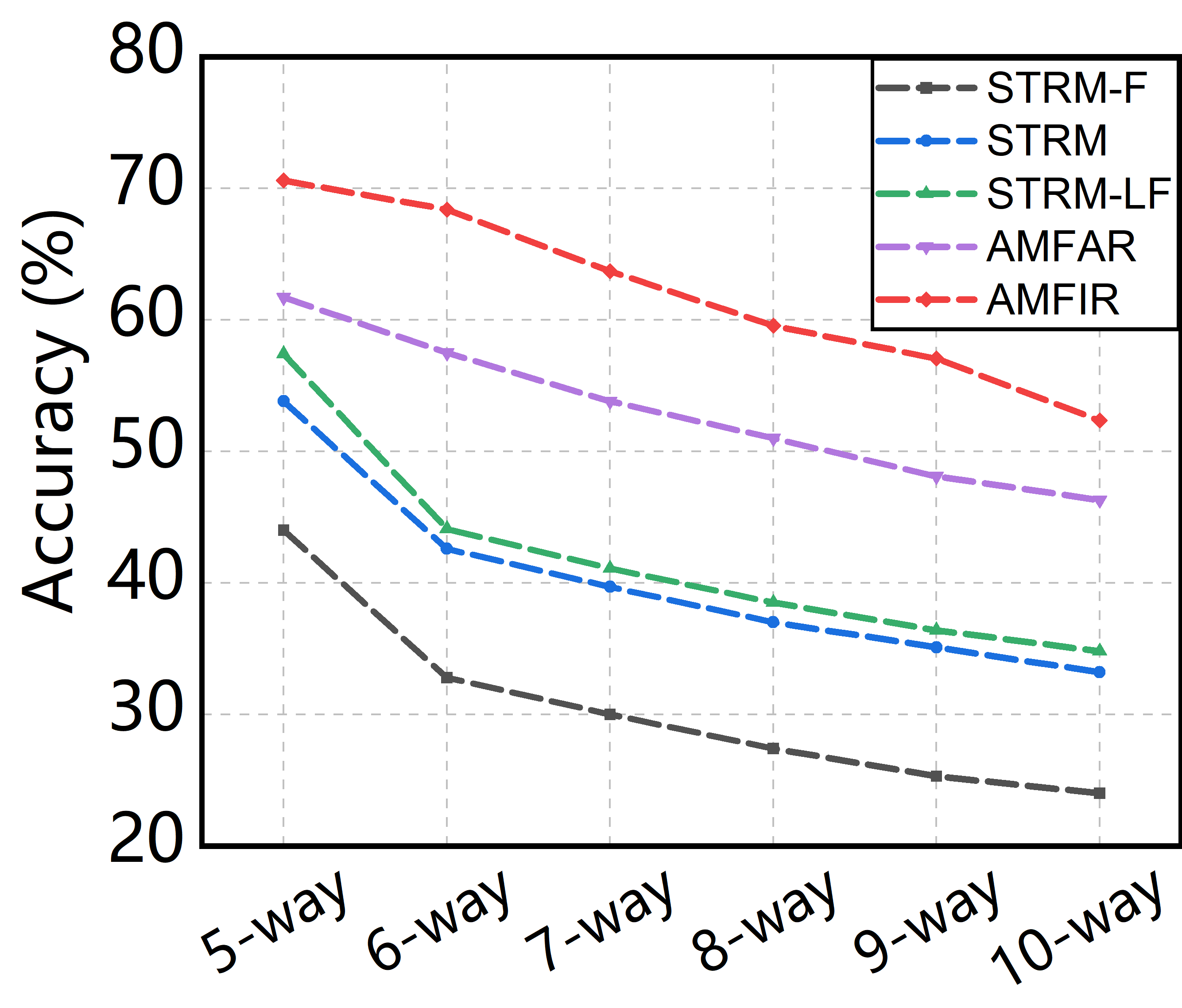}
}
\hspace{-0.1in}
\subfigure[Kinetics]{
\includegraphics[width=0.23\textwidth]{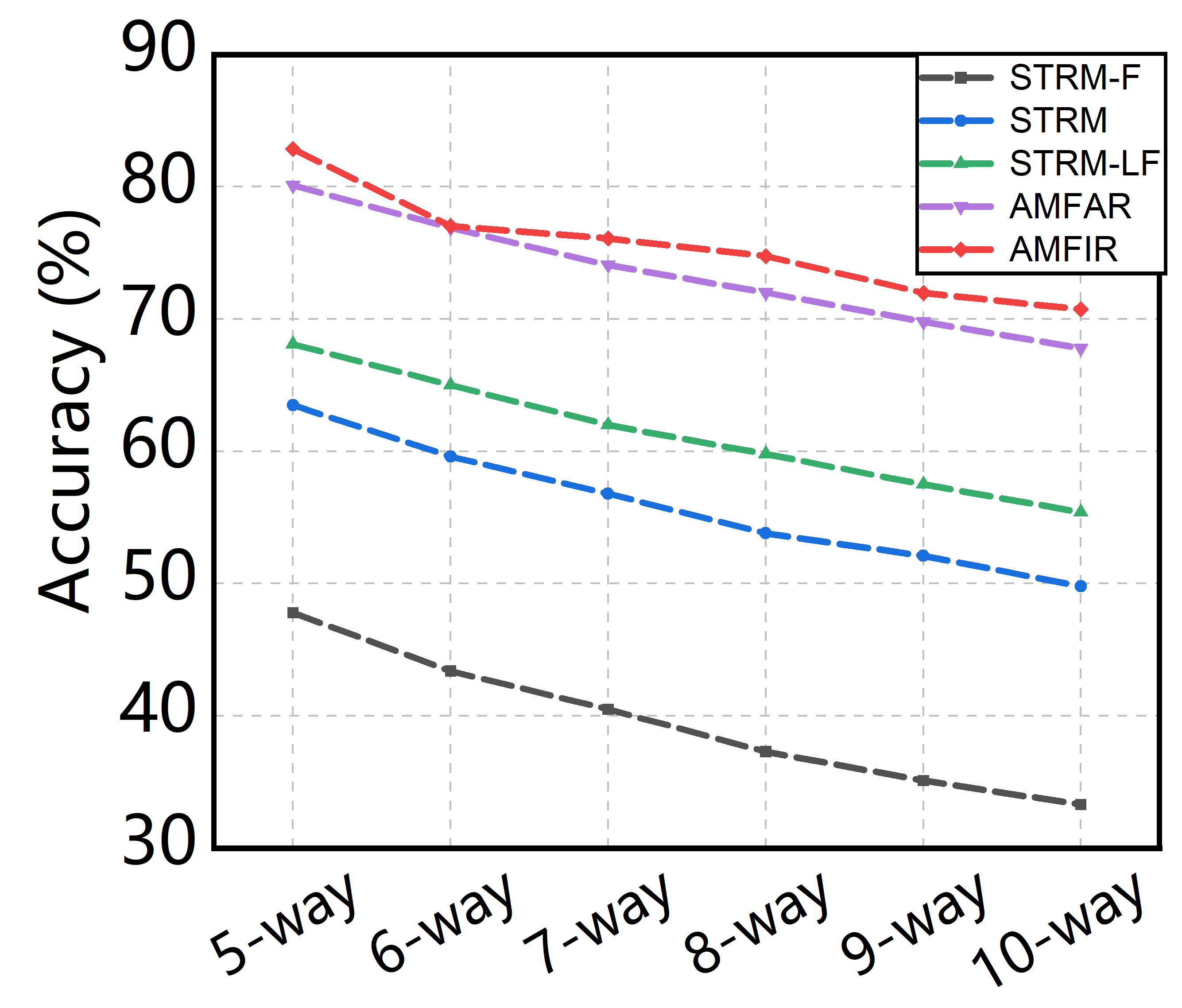}
}
\caption{N-way 1-shot performance on SSv2 and Kinetics.
}
\label{fig:N way 1 shot}
\end{figure}

\textbf{N-way Few-Shot Classification.} In the N-way Few-Shot Classification task, the AMFIR framework significantly outperforms existing methods such as STRM, STRM-F, STRM-LF, and AMFAR on the SSv2 and Kinetics datasets, achieving the best accuracy in 5-way to 10-way classification tasks. The experimental results are shown in Figure~\ref{fig:N way 1 shot}. In the 5-way task of the SSv2 dataset, AMFIR achieved an accuracy of around 70\%, while other methods were below 60\%. In the Kinetics dataset, AMFIR still achieved an accuracy of around 70\% in the 10-way task, significantly ahead of other methods. Its advantages are mainly due to the complementarity between RGB and optical flow modalities, the dynamic adaptive adjustment of the active inference module, and the enhancement of the representation of unreliable modalities by the bidirectional distillation strategy. This enables the framework to still exhibit strong robustness and generalization, even when task complexity increases, verifying its excellent performance in small-sample multi-classification tasks.

\section{Conclusion}
We have proposed the Active Multimodal Few-Shot Inference for Action Recognition (AMFIR) framework that significantly enhances few-shot action recognition by actively identifying and utilizing the most reliable modalities for each sample. By integrating active mutual distillation and adaptive multimodal inference, AMFIR effectively improves the representation learning of unreliable modalities and outperforms existing methods across multiple benchmarks. This framework highlights the potential of active inference and knowledge distillation in advancing multimodal few-shot learning through uncertainty-driven modality selection, bidirectional knowledge transfer, and context-aware fusion. Extensive experiments validate its robustness in handling sensor noise, motion ambiguity, and extreme data scarcity while maintaining computational efficiency. 

\section*{Acknowledgements}
This study is funded in part by the National Key Research and Development Plan: 2019YFB2101900, NSFC (Natural Science Foundation of China): 61602345, 62002263, 62302333, TianKai Higher Education Innovation Park Enterprise R\&D Special Project: 23YFZXYC00046, and 2024 China University Industry-Academia-Research Innovation Fund: 2024HY015.

\bibliographystyle{named}
\bibliography{ijcai25_proof_version}

\end{document}